%% file: sample-manuscript.tex
\documentclass[manuscript,acmsmall]{acmart}

\usepackage[T1]{fontenc}
\usepackage{graphicx}
\usepackage{multirow}
\usepackage{epsfig}
\usepackage{amsmath}
\usepackage{subcaption}
\usepackage{booktabs}
\usepackage{braket}
\usepackage{xcolor}
\usepackage{float}
\usepackage{wrapfig}
\usepackage{array}    % For column definitions
\usepackage{makecell} 

\usepackage{algorithm}
\usepackage{algorithmic}

\newcommand{\mj}[1]{\textcolor{black}{#1}}

\AtBeginDocument{%
  }

\setcopyright{acmlicensed}
\copyrightyear{2018}
\acmYear{2018}
\acmDOI{XXXXXXX.XXXXXXX}

\acmConference[Conference acronym 'XX]{Make sure to enter the correct
  conference title from your rights confirmation emai}{June 03--05,
  2018}{Woodstock, NY}
\acmISBN{978-1-4503-XXXX-X/18/06}

\begin{document}

\title{Mitigating Data Redundancy to Revitalize Transformer-based Long-Term Time Series Forecasting System}


\author{Mingjie Li}
\authornote{This work was done when Mingjie was with the University of Technology Sydney. The code is available at \url{github.com/mlii0117/CLMFormer}.}
\affiliation{%
  \institution{Stanford University}
  \city{Palo Alto}
  \country{United States}}
\email{lmj695@stanford.edu}
\author{Rui Liu}
\affiliation{%
  \institution{University of Technology Sydney}
  \city{Sydney}
  \country{Australia}}
\email{rliu0016@gmail.com}
\author{Guangsi Shi}
\affiliation{%
  \institution{University of Technology Sydney}
  \city{Sydney}
  \country{Australia}}
\email{guangsi.shi@monash.edu}
\author{Mingfei Han}
\affiliation{%
  \institution{University of Technology Sydney}
  \city{Sydney}
  \country{Australia}}
\email{hmf282@gmail.com}
\author{Changlin Li}
\affiliation{%
  \institution{University of Technology Sydney}
  \city{Sydney}
  \country{Australia}}
\email{changlinli.ai@gmail.com}
\author{Lina Yao}
\email{lina.yao@unsw.edu.au}
\affiliation{%
  \institution{CSIRO's Data61 and University of New South Wales}
  \city{Sydney}
  \state{New South Wales}
  \country{Australia}
}
\author{Xiaojun Chang}
\affiliation{%
  \institution{University of Science and Technology of China}
  \city{Hefei}
  \country{China}}
\email{cxj273@gmail.com}
\author{Ling Chen}
\affiliation{%
  \institution{University of Technology Sydney}
  \city{Sydney}
  \country{Australia}}
\email{ling.chen@uts.edu.au}

\renewcommand{\shortauthors}{Li et al.}

\begin{abstract}

\mj{Long-term time-series forecasting (LTSF) is fundamental to various real-world applications, where Transformer-based models have become the dominant framework due to their ability to capture long-range dependencies. However, these models often experience overfitting due to data redundancy in rolling forecasting settings, limiting their generalization ability particularly evident in longer sequences with highly similar adjacent data. In this work, we introduce CLMFormer, a novel framework that mitigates redundancy through curriculum learning and a memory-driven decoder. Specifically, we progressively introduce Bernoulli noise to the training samples, which effectively breaks the high similarity between adjacent data points. This curriculum-driven noise introduction aids the memory-driven decoder by supplying more diverse and representative training data, enhancing the decoder’s ability to model seasonal tendencies and dependencies in the time-series data. To further enhance forecasting accuracy, we introduce a memory-driven decoder. This component enables the model to capture seasonal tendencies and dependencies in the time-series data and leverages temporal relationships to facilitate the forecasting process. Extensive experiments on six real-world LTSF benchmarks show that CLMFormer consistently improves Transformer-based models by up to 30\%, demonstrating its effectiveness in long-horizon forecasting.}
\end{abstract}




\received{November 2024}
\received[revised]{Feburary 2025}
\received[revised]{March 2025}

\maketitle
\input{sec/intro}
\input{sec/relatedwork}
\input{sec/method}
\input{sec/exper}

\section{\mj{Limitations and Future Work}}
\subsection{\mj{Limitations}}
\mj{One limitation of CLMFormer is its reliance on predefined curriculum learning schedules, where noise injection follows a fixed monotonic function. This approach may not always be optimal across different datasets, as the ideal noise level may vary based on the characteristics of each dataset. Additionally, while the memory-driven decoder enhances the model’s ability to capture seasonal dependencies, its fixed update mechanism may not be flexible enough for highly dynamic environments where trends shift rapidly. Lastly, CLMFormer, like most Transformer-based models, still requires a substantial amount of training data to generalize well. Although our curriculum learning strategy helps mitigate overfitting, performance on extremely small datasets remains a challenge.}

\subsection{\mj{Future Work}}
\mj{Several promising directions can further improve CLMFormer. First, adaptive curriculum learning could replace the fixed noise injection strategy by incorporating reinforcement learning or meta-learning to dynamically adjust the dropout rate based on dataset characteristics, ensuring optimal training progression. Second, the memory-driven mechanism could be extended to include personalized memory modules, allowing for adaptive seasonal pattern modeling tailored to specific time-series sequences, making it particularly beneficial for applications such as patient-specific healthcare forecasting or financial market predictions. Third, scaling CLMFormer to large-scale foundation models could enable pretraining on diverse multi-domain time-series datasets, improving generalization and enabling cross-domain transfer learning. Addressing these challenges will further enhance the flexibility, efficiency, and adaptability of CLMFormer in real-world forecasting applications.}

\section{Conclusion}
In this paper, we propose a novel pipeline, named CLMFormer, to revitalize Transformer-based LTSF systems by mitigating data redundancy stemming from the rolling forecasting settings. We first discuss the influence of rolling forecasting settings on Transformer-based LTSF systems. Subsequently, we train a seasonal memory-driven Transformer-based model in a progressive fashion. We demonstrate that seasonal memory improves the model's ability to comprehend both multivariate and univariate sequential patterns from highly similar training samples. In addition, the progressive training schedule can effectively address overfitting issues in LTSF. We validate our methods on six public real-life benchmarks with five different prediction lengths ranging from 24 to 720-time steps. The results show that our method works exceptionally well with longer prediction sequences and can be seamlessly integrated into varying Transformer-based LTSF models and achieve up to 30\% improvement. 

%\section{Acknowledgment}
%This work is supported in part by Australian Research Council (ARC) Discovery Early Career Researcher Award (DECRA) under grant No. DE190100626.

\section*{Acknowledgements}
This work was supported by the Australian Research Council Discovery Project No. 210101347 and the Australian Research Council (ARC) Discovery Early Career Researcher Award (DECRA) under grant No. DE190100626.

\bibliographystyle{splncs04}
\bibliography{sample-base}

\end{document}

%% file: sec/intro.tex
\section{Introduction}
\mj{Long-term time series forecasting (LTSF)~\cite{acmsurvey2,ACMtimerserissurvey} stands as a cornerstone within a multitude of practical applications for complex real-life scenarios, such as finance~\cite{DBLP:journals/ijon/TangSZYHJTL22financial}, traffic, weather~\cite{bi2023accurate} and hospital admission rates prediction~\cite{DBLP:journals/midm/RamanADKCAAACIBM23}, wielding the power to shape crucial outcomes. Witness the success of Transformer~\cite{vaswani2017attention} and further applications in numerous fields~\cite{li2022video,li2023dynamic,li2024contrastive}, researchers have begun to utilize the Transformer and its evolutionary descendants~\cite{lee2024ts-fastformer,nie2022time,zhou2022fedformer,Zhou2020informer} as the de facto backbone in this intricate landscape. It showcases unparalleled prowess in handling extensive sequence data, elucidates patterns, and unveils latent structures that underscore sequential phenomena. More details, these Transformer-based LTSF systems have demonstrated strong performance, for which they can predict hundreds of time steps ahead, in contrast to other sequence processing networks and traditional methods that were mostly only capable of making predictions for 48-time points or less \cite{Hochreiter1997LongSM,QinSCCJC17,wen2018multihorizon}. Transformer-based architectures have demonstrated exceptional capability in modeling long-range dependencies, making them well-suited for LTSF~\cite{Wu2021autoformer}. }

\mj{Despite these advancements, it is important to recognize that prior LSTM-based models such as Long Short-Term Memory (LSTM)\cite{Hochreiter1997LongSM} and its variants, including Seq2Seq\cite{sutskever2014sequence} and Attention-based LSTM~\cite{bahdanau2014neural}, approached LTSF differently by leveraging their innate ability to process sequential data. These methods excelled in capturing local temporal dependencies but struggled with learning long-range dependencies due to vanishing gradient issues. LSTM-based methods also relied heavily on carefully tuned window sizes for rolling forecasting, which inadvertently introduced redundancy similar to Transformer-based systems. However, unlike Transformers, LSTM methods lacked the architectural flexibility to mitigate the overfitting caused by redundant data, as increasing network depth often led to performance degradation rather than enhancement.}

Strikingly, when playing with these Transformer-based LTSF systems~\cite{lee2024ts-fastformer,Wu2021autoformer,Zhou2020informer}, we observed that these models frequently attain their apex performance on validation datasets within a remarkably abbreviated span of training epochs, sometimes even the first epoch~\cite{Zhou2020informer}, and then the performances dropped gradually. This observation hints at the latent underexploitation of the Transformer-based systems' innate potential. The emergence of such a phenomenon can typically be attributed to a confluence of two primary factors: insufficient training data and the constrained parameter within the network architecture. However, the inherent modularity of these architectures allows for facile parameter augmentation by virtue of stacking additional Transformer modules. This architectural extensibility, coupled with the recognition that the constrained parameter space might not be the predominant bottleneck, guides our inference toward the more probable origin: the paucity of comprehensive and high-quality training data.

\mj{Existing models like Fedformer~\cite{zhou2022fedformer} and Informer~\cite{Zhou2020informer} have introduced specialized mechanisms to alleviate some limitations of redundancy in sequential data. Fedformer incorporates multi-scale decomposition to capture seasonal trends, while Informer uses a sparse self-attention mechanism to reduce computational overhead and focus on key temporal correlations. However, both approaches are still prone to the pitfalls of data redundancy stemming from rolling forecasting settings. Specifically, while sparse attention reduces the volume of redundant computations, it does not inherently address the similarity between consecutive training samples, which persists as an obstacle for longer sequences. Similarly, multi-scale decomposition focuses on improving representation but does not directly mitigate the negative impacts of overlapping inputs in rolling forecasting.}

\begin{figure}[t]
\centering
\includegraphics[width=0.5\textwidth]{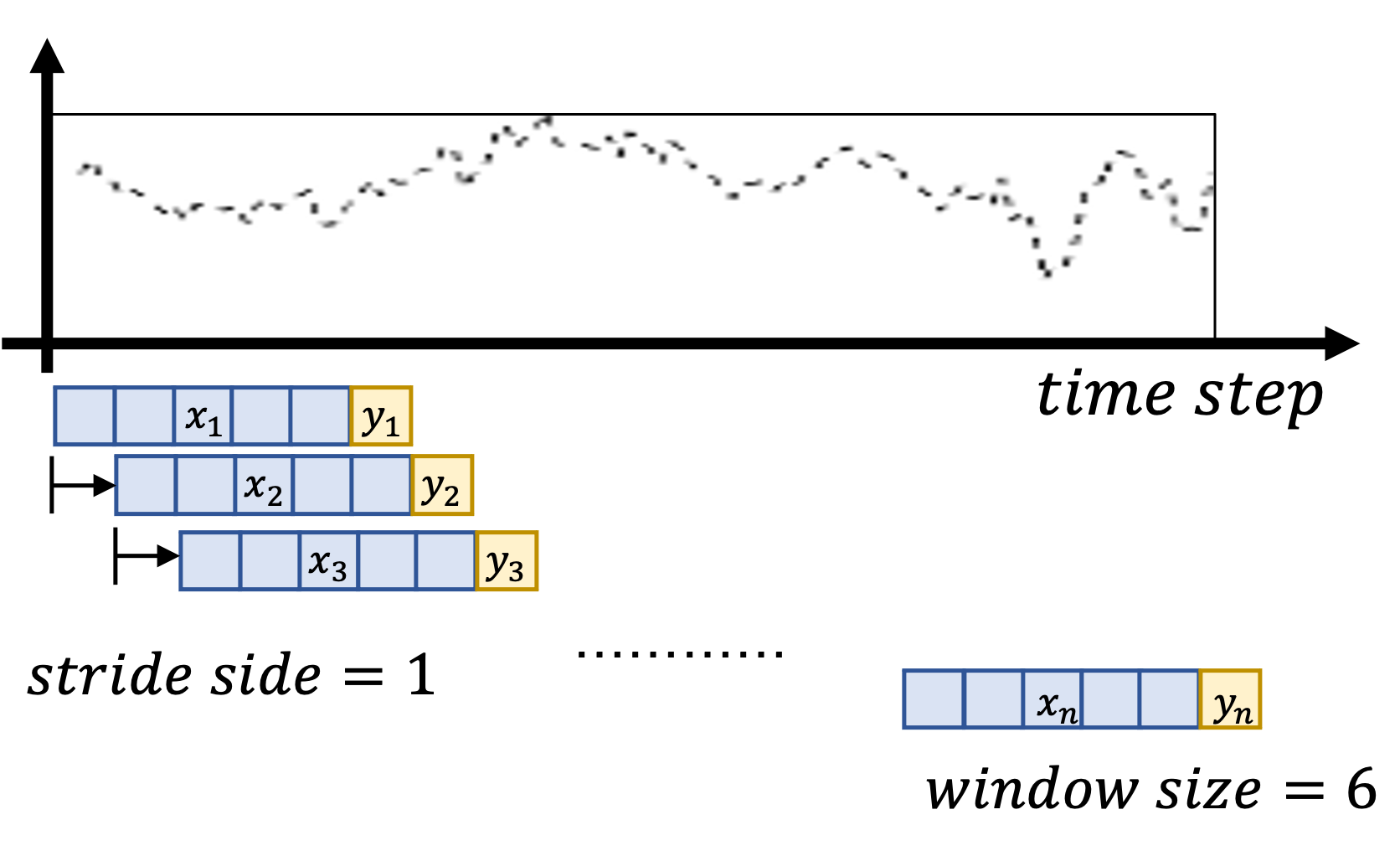}
\caption{Illustration of the rolling forecasting setting with stride size=1 and window size=6.}
\label{rolling}
\end{figure}

Delving further into this phenomenon, we found that there is severe data redundancy within the training samples, stemming from the common practice of rolling forecasting settings (see Figure.\ref{rolling} for more details) during data augmentation. Under the rolling forecasting setting, if the stride size is 1, then the previous and current training samples are only one data point apart from each other. Since making longer predictions typically requires longer input sequences, training samples for LTSF models have a higher degree of similarity. For example, when the model’s input length is 2, the similarity between two consecutive training inputs is 50\% (1/2); however, if we increase the model’s input length to 168, the similarity between two consecutive training inputs rises to 99.4\% (167/168). Therefore, the LTSF models are more likely to suffer from overfitting due to such data redundancy. \mj{This issue significantly limits the diversity of training samples, reducing the models’ ability to generalize to unseen patterns, a challenge shared with LSTM-based models that rely heavily on overlapping inputs during training.} One could argue that we could decrease the training sample similarities by increasing the stride size; however, the number of training samples decreases by the factor of the stride size, as the training sample numbers are calculated by: 

\begin{align}
  N_{t}=\frac{N_{d}-S_w+1}{S_s}\rm{.}
\end{align}
where $N_{t}$ represents the number of training samples, $N_d$ represents the total number of time points in the training data, $S_w$ refers to the window size, and $S_s$ is the stride size. Although increasing stride sizes can decrease the sample similarities, it causes a decrease in training sample numbers, which also increases the likelihood of overfitting. 

To address the aforementioned data redundancy issue, we propose CLMFormer, a novel pipeline with curriculum learning and a memory-driven decoder, designed to revitalize varying Transformer-based LTSF systems. Since data redundancy directly results from the heightened similarity present within the training data, our priority is to break this similarity while concurrently preserving the necessary training sample volume. To achieve this, we leverage the concept of curriculum learning, gradually increasing training difficulty and data variety by progressively introducing Bernoulli noise to the training data through a dynamic dropout scheme. Such a scheme reduces the propensity for overfitting while simultaneously enhancing the model's capacity to capture sequential patterns. To further enhance the Transformer’s ability to recognize intricate sequential patterns from highly similar training data, we propose a memory-driven decoder to facilitate the forecasting process. This is accomplished by adopting a seasonal memory and memory-conditioned layer normalization operation, both of which effectively utilize past time-series information and introduce additional computations to enhance the model’s capability for complex pattern recognition. The main contributions of this paper are summarized as follows:

\begin{itemize}
    \item We uncover the swift validation peak in Transformer-based LTSF systems, attributed to data redundancy stemming from rolling forecasting settings. 
    \item We propose a novel pipeline, CLMFormer, introducing curriculum learning and a memory-driven decoder to revitalize Transformer-based LTSF systems.
    \item We employ a progressive training schedule to increase the training difficulty and data variety and propose a seasonal memory matrix to recognize sequential patterns from highly similar training data.
    \item Extensive experiments are performed, and the results show that our method can be seamlessly plugged into various transformer-based LTSF models and improve their performances by maximally 30\%.
\end{itemize}

\mj{The rest of this paper is organized as follows: Section 2 provides a comprehensive review of related work in LTSF and Transformer-based architectures. Section 3 presents the CLMFormer framework, detailing its progressive curriculum learning strategy and memory-driven decoder. Section 4 describes the experimental setup, including datasets, evaluation metrics, and baseline comparisons, followed by a thorough performance analysis. Section 5 discusses the limitations and potential future improvements of CLMFormer, while Section 6 concludes the paper by summarizing key findings and highlighting the broader impact of our approach.}

%% file: sec/relatedwork.tex
\section{Related Work}

\subsection{Time Series Forecasting} 

Time series forecasting (TSF)~\cite{DBLP:journals/csur/BenidisRFWMTGBS23,TKDE3} is a critical task in various domains, encompassing domains such as finance, energy, healthcare, and more. Early approaches to TSF often revolved around traditional statistical and machine learning methods. These methods include ARIMA models~\cite{box1968somearime}, exponential smoothing methods~\cite{gardner1985exponential}, and seasonal decomposition of time series algorithms~\cite{cleveland1990stl}. These techniques have been widely used due to their simplicity and interpretability. However, they often struggle to capture intricate temporal patterns and dependencies present in real-world data.

In recent years, deep learning techniques have gained traction for TSF. RNNs, including LSTM\cite{Hochreiter1997LongSM}, GRU\cite{gru}, and DeepAR\cite{salinas2020deepar}, have demonstrated their capability to capture temporal patterns. However, these models easily suffer from error accumulation, a phenomenon where small errors generated during earlier time steps can propagate and magnify as the sequence progresses\cite{bengio2015scheduled}. This issue can significantly impact the overall accuracy and stability of the model's predictions, especially in long sequences or when dealing with complex relationships within the data. Despite these challenges, deep learning approaches have continued to evolve, showing remarkable advancements in multivariate time series forecasting~\cite{TKDECheng}. Tang \textit{et al.}~\cite{TangTKDE} introduced Enhanced-BF, a novel factorization approach for multivariate time series data, overcoming challenges like scattered data distribution through a three-phase process that improves robustness, efficiency, and accuracy in extracting meaningful features. Jin \textit{et al.}~\cite{TKDEjin} addressed limitations of discrete neural architectures, computational complexity, and reliance on static graph structures by using dynamic graph neural ordinary differential equations to improve multivariant prediction performances. Beyond these developments, innovative architectures like TimeMixer~\cite{timemixer} and TimeMixer++~\cite{wang2024timemixer++} have redefined time-series forecasting by leveraging multiscale-mixing and advanced pattern extraction techniques. TimeMixer disentangles seasonal and trend components across fine and coarse scales using Past-Decomposable-Mixing (PDM) and Future-Multipredictor-Mixing (FMM), achieving state-of-the-art performance in both short- and long-term forecasting. Building on this, TimeMixer++ introduces multi-resolution time imaging (MRTI) and multi-scale mixing (MCM) to extract comprehensive temporal and frequency-domain patterns, enabling superior adaptability across diverse tasks. However, these models require substantial data to train effectively, while real-world time-series datasets are often small, domain-specific, and strongly biased by trends, limiting their practicality and scalability in certain applications. \mj{Despite their sophistication, these models rely heavily on abundant and diverse training data. However, real-world time-series datasets often exhibit redundancy, particularly when rolling forecasting settings are applied, which undermines generalization and scalability.}

\subsection{Transformer-based LTSF Systems} 
Due to their superior performance in processing sequential data, Transformer-based models have achieved state-of-the-art results in numerous time-series tasks \cite{wen2022transformers}, particularly in long-term series forecasting (LTSF) problems \cite{cao2023inparformer,Wu2021autoformer,Zhou2020informer,jin2023expressive}. Research in this area has focused on several key directions. Some studies aim to reduce computational costs to enable longer predictions by employing sparse attention mechanisms, which reduce memory costs from O($L^{2}$) to as low as O($LlogL$), where L refers to the sequence length in attention layers \cite{Wu2020AdversarialST,Zhou2020informer}. Other works emphasize architectural improvements or enhanced patch representations to better learn seasonal and intricate temporal patterns.

Transformer-based models, however, still face challenges in sequential relationship representation and often suffer from slow training and inference speeds due to their deep encoder-decoder architectures. To address these issues, several innovative approaches have been proposed. For instance, Lee \textit{et al.}~\cite{lee2024ts-fastformer} introduced the Sub Window Tokenizer, which simplifies and compresses input sequences with the help of a pre-trained encoder. The Autoformer model \cite{Wu2021autoformer} introduced an Auto-Correlation mechanism, allowing sub-series level aggregation to effectively handle periodicity in time-series data. FEDformer \cite{zhou2022fedformer}, in contrast, adopts frequency-enhanced blocks and attention mechanisms, utilizing random subsets of frequency components along with learnable complex-number parameters to construct sparse temporal representations. InParformer \cite{cao2023inparformer} integrates evolutionary seasonal-trend decomposition modules with an interactive parallel attention mechanism, combining frequency-aware and time-aware attention to better capture meaningful patterns in time-series data. PatchTST \cite{nie2022time} improves representation learning by using subseries-level patches and promoting channel independence. Despite these advancements, Zeng \textit{et al.} \cite{zeng2023transformers} argued that Transformer-based LTSF models remain underutilized and proposed a simple linear model that surprisingly achieved superior performance. \mj{While these approaches effectively reduce model complexity and improve representation, they do not address the issue of training data redundancy caused by overlapping input sequences, particularly in rolling forecasting settings. Such redundancy limits the diversity of training samples and amplifies the risk of overfitting, which is especially problematic in long-term forecasting tasks.}

The rise of deep Transformer networks, such as foundation models and large language models (LLMs), marks a significant paradigm shift in time-series forecasting. These models, originally developed for natural language processing tasks, have demonstrated unparalleled capacity for learning from massive datasets, capturing long-term dependencies, and adapting to diverse tasks. Foundation models, like the decoder-only model for time-series forecasting proposed by Das \textit{et al.}~\cite{timesfoundationmodel}, leverage their powerful pre-training capabilities to provide a unified architecture for forecasting across various applications. Similarly, Jin \textit{et al.}~\cite{LLMintimeseriesforecasting} explored the use of existing LLMs to revolutionize time-series analysis, framing forecasting tasks as sequence generation problems and benefiting from the transfer learning capabilities of these large-scale models.

Despite their promise, employing these large models in time-series forecasting presents significant challenges. Unlike natural language tasks, time-series datasets are often small, domain-specific, and strongly influenced by underlying trends and seasonality. These characteristics make time-series data inherently biased, limiting the generalization capabilities of large models pre-trained on diverse datasets. Moreover, the massive scale of foundation models and LLMs demands extensive computational resources and vast quantities of training data, which are often unavailable in time-series domains. This data scarcity, combined with the unique and predictable nature of time-series patterns, makes training such large Transformer networks difficult and inefficient. Consequently, while these models hold great potential, their practical application in time-series forecasting must address the trade-offs between model complexity, data requirements, and the specialized characteristics of time-series data. In this paper, we revisit these models and uncover the issue attributed to data redundancy stemming from rolling forecasting settings.  \mj{Most importantly, data redundancy caused by rolling forecasting settings persists as a key bottleneck in long-term forecasting systems, compounding challenges associated with small, domain-specific datasets. This redundancy not only exacerbates overfitting but also diminishes the representational quality of training data, creating a significant gap in existing Transformer-based LTSF solutions. To bridge this gap, CLMFormer directly tackles data redundancy by employing curriculum learning to diversify training samples and a memory-driven decoder to enhance seasonal and dependency modeling, ensuring robust performance on long-term time-series forecasting tasks.}

%% file: sec/method.tex
\section{Methodology}
\begin{figure}[t]
\centering
\includegraphics[width=0.9\textwidth]{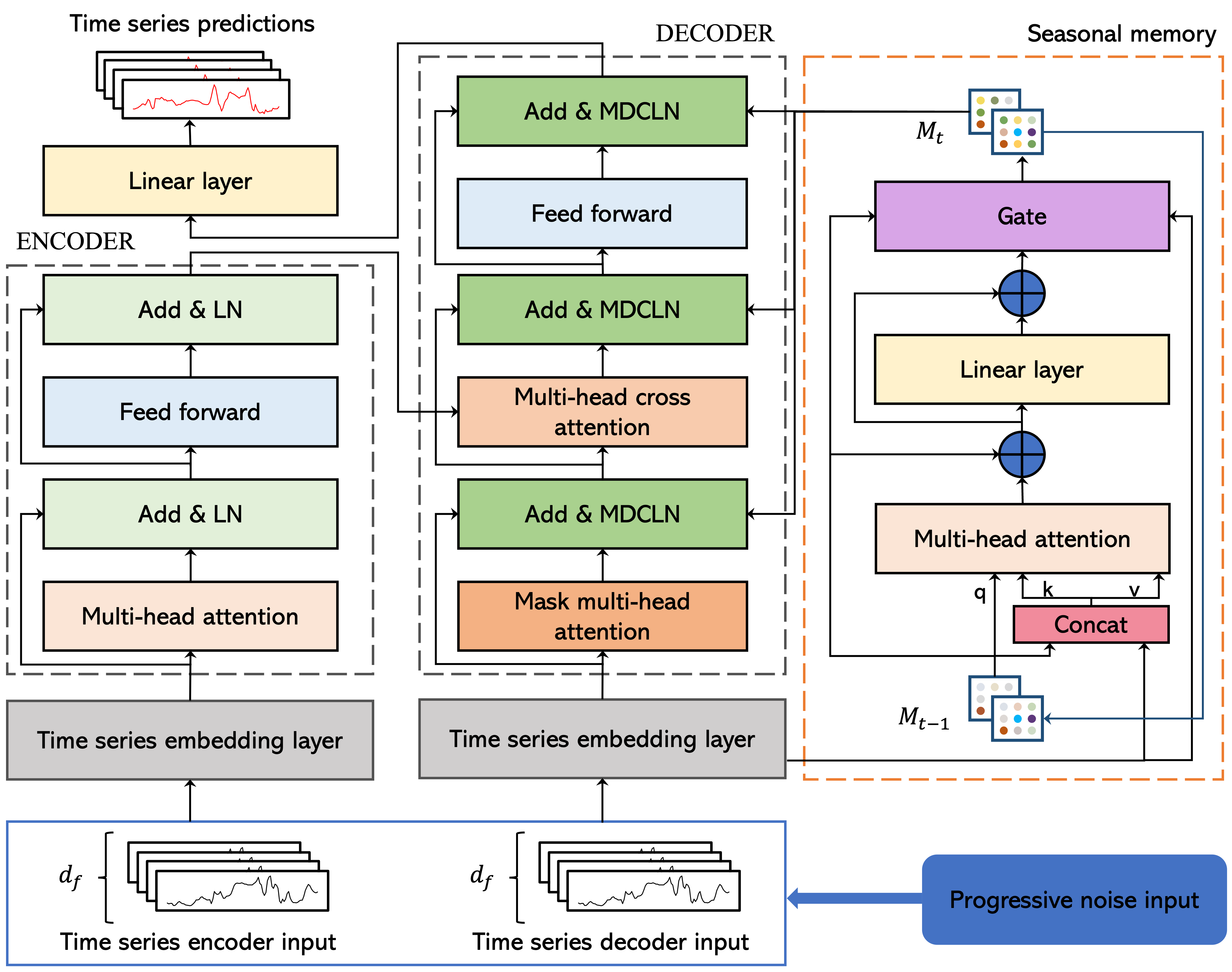}
%\vspace{-0.6cm}
\caption{An overview of the model architecture. $M_{t-1}$ represents the memory matrix from the previous prediction, $M_{t}$ represents the updated memory matrix during the current sequence prediction, which in turn is recycled to be the $M_{t-1}$ for the next prediction.}
%\vspace{-0.5cm}
\label{model}
\end{figure}

In this section, we present our solutions to revitalize transformer-based systems that specifically target the challenges faced by LTSF. Our approach is based upon the Transformer architecture; we aim to enhance Transformer models on LTSF problems by making improvements to mitigate data redundancy from both training optimization prospects and the architecture. The overview of our approach is illustrated in Figure.\ref{model}. Our model comprises three major elements, namely the input embedding, the main Transformer-based model, and the seasonal memory component, whereby we propose to train the model based on a progressive training strategy. We first introduce the background of employing Transformer models for LTSF and then describe the training strategy. The seasonal memory component is described in the remaining subsections.

Typical Transformer-based LTSF models normally train under rolling forecasting settings with a fixed window and a stride size = 1, as shown in Figure.\ref{rolling}. The input data at time t can be written as $X^{t}=\{x_{1}^{t},x_{2}^{t}...x^{t}_{\emph{L}_{S}}|x_{k}^{t} \in \rm{R}^{\emph{d}_{\emph{f}}}\}$, where ${L}_{S}$ is the input length and the feature dimension $d_{f}>1$. Moreover, the model output is $Y^{t}=\{y_{1}^{t},y_{2}^{t}...y^{t}_{\emph{L}_{p}}\}$, where ${L}_{p}$ is the prediction length, and the output feature dimension is flexible and is not limited to single sequence predictions. \mj{Algorithm.\ref{alg:clmformer} presents the training process of the whole system.}

\begin{algorithm}[ht]
\caption{\mj{Training Process of Our CLMformer}}
\label{alg:clmformer}
\begin{algorithmic}[1]
\REQUIRE Time-series input data $X$, maximum dropout rate $\theta_{max}$, learning rate $\alpha$, hyperparameter $\gamma$
\ENSURE Trained model parameters $\Theta$

\STATE \textbf{Initialize:} Memory matrix $M_0$, model parameters $\Theta$
\FOR {epoch = 1 to $N_{\text{epochs}}$}
    \FOR {batch $B$ in training set}
        \STATE Extract input sequence $X_B$ and target sequence $Y_B$
        
        \STATE \textbf{Curriculum Learning Strategy:}
        \STATE Compute dynamic dropout rate
        \STATE Apply dropout to input sequence: $X_B \leftarrow \text{Dropout}(X_B, \theta_d(t))$
        
        \STATE \textbf{Encoder Processing:}
        \STATE Encode $X_B$ using Transformer encoder to obtain hidden representation $H$
        
        \STATE \textbf{Memory-Driven Decoder Processing:}
        \FOR {time step $t$ in decoder}
            \STATE Compute attention-weighted memory update
            \STATE Apply Memory-Driven Conditional Layer Normalization
            \STATE Compute final decoder output $\hat{Y}_t$
        \ENDFOR
        
        \STATE Compute loss $\mathcal{L}(Y_B, \hat{Y}_B)$
        \STATE Update model parameters: $\Theta \leftarrow \Theta - \alpha \nabla_\Theta \mathcal{L}$
    \ENDFOR
\ENDFOR
\STATE \textbf{Return} trained model parameters $\Theta$
\end{algorithmic}
\end{algorithm}

\subsection{Base Model}\label{basemodel}

\subsubsection{Input Embedding} The input data are first passed through an embedding. Different from other types of sequential data, in time series problems, the model should capture the time information. This is extremely crucial in LSTF settings, as models are required to make predictions up to weeks in advance. We decided to utilize the time-series-specific embedding method proposed by \cite{Zhou2020informer}. The method applies a combination of the context vector, the positional embedding, and the seasonal embedding. The context vector $u^{t} \in R^{L_{S}*d_{model}}$ is obtained by projecting the model input $X^{t} \in R^{L_{S}*d_{f}}$ into a dimension of $d_{model}$ by an 1-D convolutional
layer. The goal of the context vector is to transform the input data into an appropriate dimension. Positional embedding (PE(pos)) and seasonal embedding (SE(pos)) are used to capture the local context and global time features. The formula for the input embedding is: 

\begin{align}
X_{embed}^{t}=\delta u^{t}+\rm{PE}(\rm{pos})+\sum_{k}[\rm{SE(pos)}]_{k}
\end{align}
Where $X_{embed}^{t} \in R^{L_{S}*d_{model}}$ is the embedded model input, $\delta$ is the weight factor to balance between the embeddings and the context vector and k represents the levels of the seasonal components, \textit{i.e.} days, weeks, seasons, holidays. 

\subsubsection{Encoder-Decoder Architecture} The backbone of our method lies in Transformer-based models following an encoder-decoder structure. Despite some minor differences between different variants, their general purpose is the same. The encoder takes past input sequences into the model and generates hidden states as outputs that contain time pattern information, which the decoder uses to make predictions. We decided to use generative-style decoding~\cite{Zhou2020informer}, which has been proven to achieve superior performances in LSTF problems and have exponentially better computational efficiency than conventional dynamic decoding~\cite{devlin2018bert}. The generative-style decoding generates all predictions in one step, with the decoder feeding vector $X^{t}_{feed,dec}$ presented as:

\begin{align}
X^{t}_{feed,dec}=\rm{Concat}\it{(X_{dec}^t,X_0^t)}\in \it{{R^{(L_{dec}+L_{p}) * d_{f}}}},
\end{align}
where $X_{dec}^t \in R^{L_{dec} * d_{f}}$ is the time-series input sequence for the decoder, $L_{dec}$ is the length of the decoder time-series input, $X_0^t \in R^{L_{p} * d_{f}}$  is the placeholder for the prediction sequence, and the $L_{p}$ is the prediction length.

\subsubsection{Loss Function} The loss function we use is Mean Squared Error (MSE) loss on the target sequences, and it is then backpropagated across the entire network. 
\begin{align}
L_{MSE}=\frac{1}{n}\sum_{i=1}^{n}(y_i-\hat{y_i})^2
\end{align}
where $y_i$ is the true value for the $i^{th}$ time step in the output sequence, $\hat{y_i}$ is the predicted value, and n is the total number of values. Both $y_i$ and $\hat{y_i}$ have a dimension of $\geq{1}$.

\subsection{Progressive Training Strategy}
As mentioned before, we argued \mj{traditional Transformer-based models, the strong inductive bias enables them to effectively capture complex temporal dependencies. However, this also means that these models require large-scale datasets for training to generalize well. Unfortunately, most existing TSF benchmarks contain relatively simple and repetitive patterns, which leads to fast convergence and severe overfitting in these Transformer-based models.} Therefore, we decided to break these similarities during the training process but kept the rolling forecasting settings. Because it does augment the training samples effectively and efficiently. Since increasing stride size for the rolling window does not help with the overfitting issue, inspired by the masking strategy~\cite{devlin2018bert}, we decided to mitigate the data redundancy issue by adding sample variety. These noises should be added in a manner such that the rate is slower at the beginning to ensure that the model has a good grasp of the data pattern before we start to increase the rate of introducing noises to prevent the model from overfitting.  To achieve this goal, we decided to alter dropout rates, as there is a proven connection between dropout and noises \cite{Bishop1995TrainingWN,Wager2013DropoutTA,Zhai2015DropoutTO}. We dynamically alter the dropout rate during training based on a monotone function proposed by \cite{Morerio2017CurriculumD}, whereby the function rate starts slower and later increases as training proceeds. The dropout rate function is based on:
\begin{align}
\theta_d(t)=\min(\theta_{max},1-\theta_{max}-{(1-\theta}_{max})\exp(-\gamma t))
\end{align}
where t is the $t^{th}$ number of 100 iterations, $\theta_d$ is the dropout rate, $\theta_{max}$ is the maximum dropout rate,  and $\gamma$ is a hyper-parameter typically set between 0.001 to 0.01. \mj{Compared with other scheduling methods such as linear or cosine because it offers a gradual and adaptive transition in the dropout rate. The linear scheduling increases the dropout rate at a constant pace, which can lead to a too-aggressive or too-slow regularization effect at different training stages. Cosine scheduling is commonly used for learning rate decay but is less effective for dropout scheduling, as it introduces oscillations that may interfere with stable training progression.}

\begin{figure}
    \centering
    \includegraphics[width=0.75\linewidth]{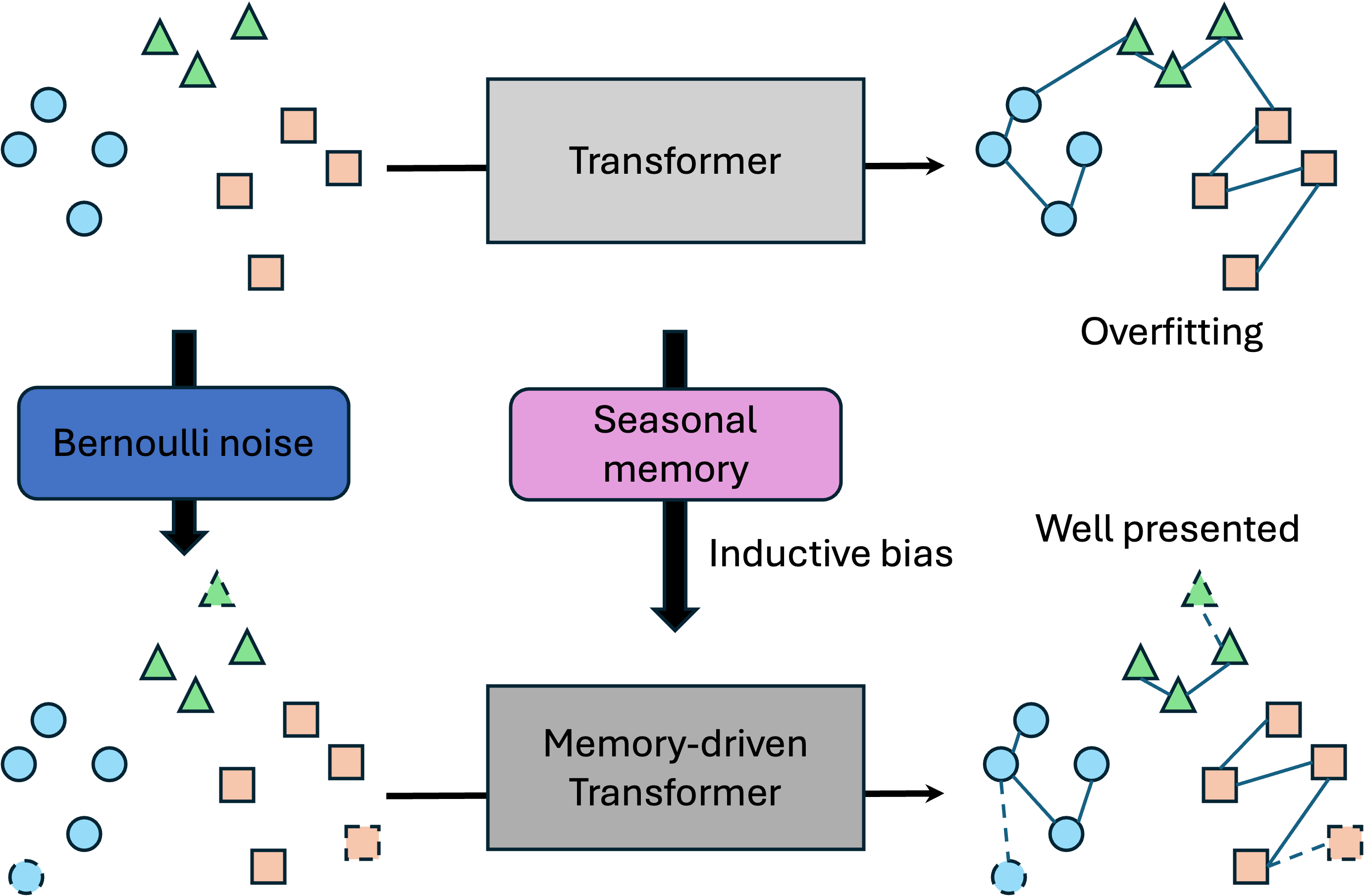}
     \caption{The interaction between curriculum learning and memory-driven mechanism.}
    \label{correlation}
\end{figure}

\subsection{Seasonal Memory-driven Forecasting}
We illustrate the interaction between curriculum learning and our memory-driven mechanism in Figure.\ref{correlation}. \mj{While injecting noises helps mitigate overfitting, it also introduces a new challenge—without proper inductive biases, the model may misinterpret noise as meaningful information, leading to instability in predictions. To counteract this, we introduce a memory-driven mechanism, which serves to reinforce the model’s intrinsic inductive bias. This module ensures that while the model is exposed to progressively noisier training samples, it does not become overly sensitive to noisy data points.} Applications of the memory concept have demonstrated superior performances in a variety of sequential tasks \cite{vqamemory2018,Ma2020MemoryAG,Fei21a}. In particular, \cite{Xu2020TensorizedLW} demonstrated that the memory concept is effective in improving LSTM’s ability on multivariate TSF. Therefore, we argued that this concept should also work on Transformer models under the multivariate forecasting setting. 

There are various ways of implementing memory into Transformers \cite{jiang2019web,BaninoBKCZHBBKB20,chen2020raio,memoryma2021,DBLP:conf/nips/ZhouMWW0YY022}. Our goal is to improve the Transformer's ability to handle the increased data complexity that is brought about and to recognize hidden sequential patterns from highly similar training data. We decided to incorporate a memory unit into the decoder block because the decoder block is highly relevant to the output predictions. 
To achieve this goal, we proposed a seasonal memory, having a self-attention layer that can help to further analyze the hidden sequential patterns that are missed by the encoder-decoder component by combining information stored in the memory matrix together with the model input. The way to connect the seasonal memory block to the decoder is via a memory-driven conditional layer normalization (MDCLN) layer, the general formula for this connection can be written as:

\begin{align}
   \rm{MDCLN(Att(h_{l-1}),M_{t})}
\end{align}
where MDCLN(.) is the memory-driven conditional layer normalization, Att(.) represents the attention layer in the decoder block, h represents the output state from the previous layer $l - 1$, and $M_{t}$ refers to the updated memory matrix at time point $t$.

\subsubsection{Memory-driven Conditional Layer Normalization} 
\begin{figure}[t]
\centering
\includegraphics[width=0.9\textwidth]{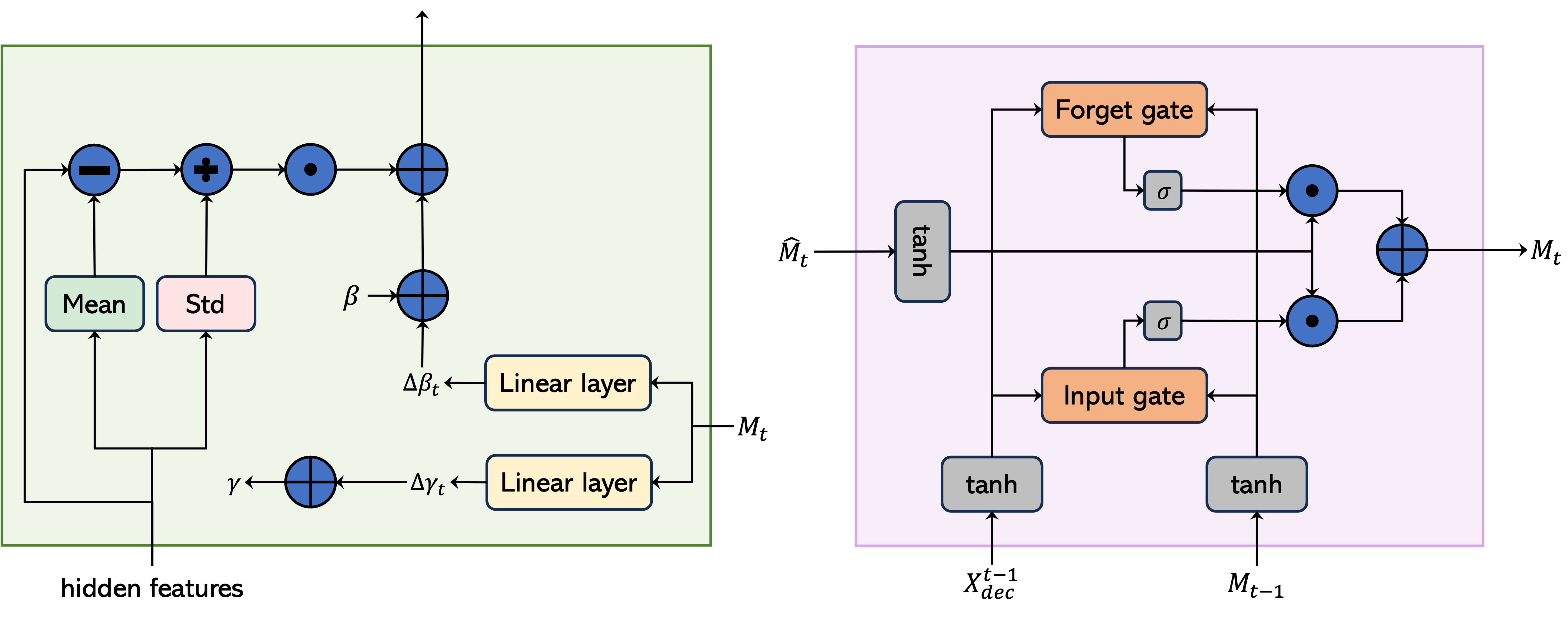}
\caption{Detailed architectures of the memory-driven conditional layer normalization layer (left, green box) and the gate mechanism (right, purple box).}\label{gate}
\end{figure}

The Memory-driven Conditional Layer Normalization (MDCLN) is a way to integrate seasonal memory into the decoder, where outputs from the decoder’s self-attention layer undergo layer normalization along with the memory information provided from the seasonal memory block. We present its detailed architecture in Figure.\ref{gate} right. The reason for using layer normalization instead of batch normalization is that input data are normalized before feeding into models as common practice for time-series forecasting. Hence, instead of normalizing values among batches, we use layer normalization to normalize the features within each batch to give each time feature an equal weight for generating predictions.

In MDCLN, we update $\gamma$ and $\beta$ based on the memory matrix $M_t$, as they are the two most essential parameters for normalization:
\begin{align}
\gamma_t=\gamma+\it{f_{l}(M_{t})},
\end{align}
\begin{align}
    \beta_t=\beta+f_{l}(M_{t}).
\end{align}
Where $f_{l}(.)$ represents the linear layer. The MDCLN is achieved by:

\begin{align}
\rm{MDCLN}(\it{h_{l-1}},\it{M_{t}})= \gamma_t \odot \frac{h_{l-1}-\mu}{\sigma}+\beta_t+f_{l}).
\end{align}
Where $\mu$ and $\sigma$ are the mean and standard deviation of $h_{l-1}$, and $h_{l-1}$ is the output of self-attention at layer $l-1$ in the decoder.

\subsubsection{Seasonal Memory Matrix} In addition to the further calculations that the seasonal memory unit can bring, another main characteristic of this unit is that it can read/write information from/to a memory matrix which assists with analyzing time patterns. In NLP problems, we use dynamic decoding \cite{devlin2018bert}; the memory matrix works by maintaining information from the prediction of the last token/word while predicting a paragraph, and this matrix is re-initialized after each paragraph, as each paragraph is perceived as being independent. However, as explained above, we use generative-style decoding for our task, in which it generates all predictions in a single step \cite{Zhou2020informer}. In this case, the traditional matrix initialization method will not work, as we do not generate individual time point predictions one by one. In addition, for each forecasting problem, the input sequences for making previous predictions are still helpful for any future prediction, as they come from the same time series and share similar pattern interactions. Therefore, we maintain a memory matrix for each specific forecasting task, where we keep the matrix output from the seasonal memory at the prediction time point $t-1$ as the input matrix at time point $t$; this matrix is updated after every prediction by a gating mechanism to control the ratio of the memory matrix being updated. Since more recent model inputs are weighted higher than those from the distant past, the matrix can maintain all relevant time information. In addition, this matrix will continue updating even after training, which ultimately increases the model’s generalization power. The seasonal memory unit updates the memory matrix by first calculating the self-attention based on the matrix $M$ at time point $t-1$,  where $\rm{Q}=\it{M_{t-1}}\rm{\times\ W_q}$, $\rm{K}=\rm{[\it{M_{t-1}};\ embed(X_{feed,dec}^t)] \times\ \rm{W_k}}$ and $\rm{ V=[\it{M_{t-1}};\ embed(X_{feed,dec}^t)]} \times\ \rm{W_v}$. The $\rm{W_k}$, $\rm{W_v}$ and $\rm{W_q}$ are trainable weights, and $[M_{t-1};embed(X_{feed,dec}^t)]$ is the row-wise concatenations between $M_{t-1}$ and the embedded decoder feeding vector. We then calculate the self-attention score $Z$ from Q, K and V. After the self-attention layer, we go through a feedforward layer to produce $\bar{M_t}$, which is calculated as:
\begin{align}
\bar{M_t}= \rm{Feedforward}\it{(Z+M_{t-1}) + Z + M_{t-1}}
\end{align}
$\bar{M_t}$ is then passed through a gating mechanism to calculate the ratio of the $\bar{M_t}$ to be added to the final $M_t$:
\begin{align}
M_t=\sigma\left(G_f^t\right)\ \odot\ M_{t-1}+\sigma\left(G_i^t\right)\ \odot\ \bar{M_t}\rm{,} 
\end{align}
where $G_i^t$ is the input gate, $G_f^t$ is the forget gate, $\odot$ is the Schur product, and $\sigma(.)$ is the sigmoid function.

\subsubsection{Gate Mechanism} \mj{The gate mechanism decides the proportion of new information from $\bar{M_t}$ to incorporate and the proportion of $M_{t-1}$ to discard to produce the final $M_{t}$. As shown in Figure.\ref{gate} left, the gate mechanism has two key components: the input gate $G_i^t$ and the forget gate $G_f^t$. The two gates are calculated by balancing the embedded decoder input $X_{feed,dec}^{t-1}$ and the previous memory matrix at $t-1$ ($M_{t-1}$), where we duplicate the $embed(X_{feed,dec}^{t-1})$ to match the dimension of $M_{t-1}$. These two gates have the following formulas:
\begin{align}
G_i^t= embed(X_{feed,dec}^{t-1}) \times W_i+\tanh{\left(\it{M_{t-1}}\right)}\times\ \it{U_i},
\end{align}
\begin{align}
G_f^t= embed(X_{feed,dec}^{t-1}) \times W_f + \tanh{\left(\it{M_{t-1}}\right)}\times\ \it{U_f}.
\end{align} 
Where $W_i$, $W_f$, $U_i$, and $U_f$ are trainable weights for the two gates.}

%% file: sec/exper.tex
\section{Experiments}

\subsection{Datasets}

We conduct extensive experiments on 6 public real-life benchmark datasets. For all datasets, train/valid/test sets are split following the ratio of 0.7/0.1/0.2 in chronological order. \textbf{ETTh} (Electricity Transformer Temperature) \cite{Zhou2020informer}. The dataset contains information about six power load features and the target value ``oil temperature''. We use the two hourly datasets from the source named $\rm{Etth_{1}}$ and $\rm{Etth_{2}}$. We use the first 20 months of data, \mj{which consists of approximately 17,400 hourly samples per dataset. The six features include electrical load-related attributes such as power factor, active power, and reactive power;} \textbf{Weather}\footnote{https://www.ncei.noaa.gov/data/local-climatological-data}. The dataset contains hourly data about climate information from around 1,600 locations in the USA between 2018 and 2020. The data consist of seven quantitative climatological features with ``dry bulb temperature'' as the target.  \mj{Each location contributes approximately 26,300 samples, and the features include variables such as wind speed, humidity, and atmospheric pressure;} \textbf{Air Quality}\footnote{https://archive.ics.uci.edu/ml/datasets/Beijing+Multi-Site+\\Air-Quality+Data}. The dataset consists of eleven quantitative hourly features relevant to air quality with the target value set as ``PM10'' from 12 national air quality monitoring sites. We use the data from the ``Wanshouxigong'' station. \mj{This dataset contains approximately 35,000 hourly samples, and features include pollutants such as NO$_2$, CO, SO$_2$, and ozone concentrations, in addition to meteorological factors like temperature and wind speed;}
\textbf{Traffic}\footnote{https://archive.ics.uci.edu/ml/datasets/Metro+Interstate+\\ Traffic+Volume}. The dataset contains the traffic condition data measured between Minneapolis and St Paul, Minnesota, USA. The dataset has five hourly numerical features with the target value of ``traffic volume". \mj{It consists of around 48,000 hourly samples, and features include traffic speed, lane occupancy, and weather-related variables;} \textbf{Exchange}\footnote{https://github.com/laiguokun/multivariate-time-series-data/blob/master/exchange\_rate}. The collection of the daily exchange rates of eight foreign countries including Australia, British, Canada, Switzerland, China, Japan, New Zealand and Singapore ranging from 1990 to 2016. \mj{This dataset comprises approximately 9,500 daily samples, with each feature representing the exchange rate of a different currency against the US dollar.}

\subsection{Experimental Details}

\subsubsection{Baselines} We use four Transformer variants as our baseline with generative-style decoding to generate the output sequences in one step \cite{Zhou2020informer}. These models are FEDformer\cite{zhou2022fedformer}, AutoFormer\cite{Wu2021autoformer}, Informer\cite{Zhou2020informer} and the vanilla Transformer\cite{vaswani2017attention}, respectively.In addition, since traditional methods like ARIMA and RNNs-based networks lack competitive performance\cite{cao2023inparformer,zhou2022fedformer}, we only compared our model's performances with the SOTA Transformer-based model, InParformer~\cite{cao2023inparformer}.

%% key parameters for Transformer backbone
\input{table/tab2}

\subsubsection{Implementation Details} Our model is constructed based on Pytorch with Python 3.7. We use a single NVIDIA GeForce RTX 3080Ti GPU for training. All of the input data undergo normalization before being fed into the model. The training is performed under the rolling forecasting setting (stride size=1). The encoder and decoder input lengths are $\{48,96,168,168,336\}$ and $\{48,48,168,168,336\}$, which correspond to the prediction lengths of  $\{24,48,168,336,720\}$, respectively. For the progressive training scheme, we update the dropout rate for every 100 iterations with the upper dropout limit of 0.1. Moreover, we apply an early stopping technique with patience=3 during training and the training epoch is set to be 10. The learning rate starts from 0.0001 and it halves its value for every epoch after the first two epochs.

\subsubsection{Model Reproducibility} To ensure model reproducibility, we present detailed parameter settings in this section. Table.\ref{trans1} describes the details for all Transformer-based models. We use the same setting for all three Transformer-based models in our experiment because these three models share a very similar structure and show optimal performance under similar settings. However, we use different settings for different prediction output lengths to adjust the model for different prediction tasks. Table.\ref{trans1} specifies the model dimension ($d_{model}$), the dimension of positional feedforward layers inside the encoder and decoder block ($d_{ff}$), and the number of attention heads in each attention layer (n\_heads). In addition, we include information about batch size, layer activation function, the encoder layer number (Enc. Layer no.), and the decoder layer number (Dec.Layer no.). Table.\ref{rm1} describes the parameters for the relational memory block. There are three key parameters for the relational memory unit: the unit dimension ($d_{rm}$), the number of memory units (Mem.slot.no), and the number of attention heads in the relational memory (Mem.head.no). Here,
we apply different relational memory settings for the three Transformer-based models as they require different settings to achieve optimal performance. 

We use two evaluation metrics to measure our model performance: Mean Squared Error (MSE) and Mean Absolute Error (MAE). \mj{MSE and MAE are among the most commonly used metrics for TSF due to their ability to measure prediction accuracy effectively. MAE measures the absolute difference between predicted and actual values. It is robust to outliers, as all errors contribute equally to the final value. MSE penalizes larger errors more heavily, making it sensitive to large deviations. It ensures that the model prioritizes reducing larger errors, which is important in forecasting tasks where extreme deviations can be critical.} \mj{We conduct paired t-tests to compare our model's performance against baselines. The p-values from these tests help determine whether the observed improvements are statistically significant. To clearly indicate statistical significance in our results, we use (\(*\)) in our tables to denote significant differences ($p<0.01$).}

%% he multivariate results co
\input{table/tab3}

\subsection{Multivariate Results}

Table.\ref{baselinetable} summarizes the multivariate results of the Transformer-based LTSF models and our proposed methods. We applied the same setting for each transformer-based model pair to achieve a fair comparison. Based on the results in Table.\ref{baselinetable}, we find that: 
(1) Our model can significantly improve the Transformer-based LTSF models' performances up to 30\% on multivariate LTSF problems across most cases; it improves the MSE and MAE scores in 20 of 25 prediction tasks in FEDformer, 23 of 25 prediction tasks in Informer and 19 of 25 tasks in vanilla Transformer. Our results show that our proposed CLMFormer with the progressive training schedule and memory-driven decoder can help Transformers with multivariate LTSF problems. (2) Our method works particularly well in longer prediction settings. For instance, when the prediction length is 720, our approach improves the Informer’s MSE score by 21.63\% for the Etth1 dataset and 20.00\% for the traffic dataset, in contrast to the 10.00\% and 14.96\% improvements, respectively, for the prediction length of 24, and this trend is consistently observed across all different models and datasets. The improved performance with longer prediction settings makes sense, considering that the seasonal memory unit is highly effective at dealing with highly similar data, and our progressive training strategy effectively addresses the overfitting issues caused by increased prediction lengths. (3) Our method can further improve the performances of the SOTA LTSF model, FEDformer, on almost all tasks. Such experimental results prove that our method can be seamlessly plugged into varying Transformer-based LTSF systems even when these models are designed to explore the long dependencies and seasonal-trend features to improve the prediction capabilities. The frequency modules in FEDformer can be considered as another kind of memory-driven paradigm, and it is encouraging to observe that our methods bring out additional improvements.

%% The univariate results comparison of our meth
\input{table/tab1}

%% blation study for the decomposition of our approach, where p.t. is the progressive training scheme and mem refers to our memory approach (three repetitions)
\input{table/tab4}

\subsection{Univariate Results} 

In Table.\ref{baselinetable-uni}, an extensive univariate results comparison is presented, comparing our method with various Transformer-based LTSF models. Notably, standout performers are indicated by values displayed in bold and red within each Transformer-based LTSF model pair, as well as in contrast to the InParformer model. Since InParformer is not an open-source work, we directly listed the numbers from the original paper\cite{cao2023inparformer} for reference. The observations yield several important insights: (1) Our method consistently outperforms alternative Transformer-based models, including FEDformer and AutoFormer, across a spectrum of prediction lengths and datasets, especially in lengthier predictions. This is evidenced by reductions in both MSE and MAE. For instance, in the ``Exchange" dataset with a prediction length of 720, our approach built upon FEDformer achieves an MSE of 1.126 compared to FEDformer's 1.301, while InParformer records 1.172. (2) Our methodology demonstrates pronounced efficacy in extended prediction scenarios, yielding substantial performance gains relative to longer prediction lengths. This pattern is consistently observed across multiple models and datasets. For instance, on the "ETTh2" dataset with a prediction length of 720, our approach improves an MSE of 0.267 in comparison to FEDformer's 0.278, and an MAE of 0.398 in comparison to AutoFormer's 0.409, which also achieves competitive performances to InParformer. (3) The integration of our approach produces enhancements when embedded within the state-of-the-art LTSF model FEDformer and AutoFormer for univariate LTSF, thereby affirming its capability to augment performance in various Transformer-based systems designed to capture extended dependencies and seasonal-trend features. This incorporation leads to additional improvements, complementing existing mechanisms to capture intricate temporal patterns and long dependencies in these models. It is also encouraging to observe the compelling superiority of our proposed methodology in addressing univariate LTSF tasks across diverse contexts. 

%% blation study of different memory matrix initialization methods (three repetitions)
\input{table/tab5}

\subsection{Ablation Study} 

The ablation study is detailed in the Table.\ref{ablation1} examines the impact of various components of the proposed model on time series forecasting accuracy across three datasets: Etth1, Air Quality, and Traffic. This study specifically looks at the performance of the Informer and FedFormer model and their enhancements through progressive training (p.t.), our memory approach (mem), and a combination of both referred to as ``ours." From the results, we can draw several key observations. Both the progressive training strategy and the memory approach independently improve the Informer's forecasting accuracy. For instance, in the Etth1 dataset, applying progressive training reduced the MSE from 1.110 to 1.021 for a 168-hour prediction horizon, and incorporating the memory approach further reduced it to 1.009. Similar improvements are seen in MAE metrics and across other datasets.
The combination of progressive training and the memory approach under``Informer + ours" consistently yields the best results. For example, it achieves the lowest MSE of 0.495 and MAE of 0.493 for a 24-step forecast on the Etth1 dataset. This pattern holds across all datasets and prediction horizons, indicating a synergistic effect when both enhancements are applied together. The improvements are more pronounced for longer prediction horizons. For instance, in the Traffic dataset, the ''Informer + ours" configuration reduces the MSE to 0.405 and the MAE to 0.300 for a 168-step forecast, suggesting that the proposed components are particularly effective at addressing challenges in long-term sequence forecasting (LTSF). While the Informer model and its enhancements focus on incremental improvements, the FedFormer model, both alone and with added enhancements (progressive training, memory approach, and combined), shows varied performance across datasets. ``FedFormer+ours'' achieves better MSE and MAE scores in the shorter and longer prediction horizons for the Etth1 dataset, and always outperforms the base model, especially in the Air Quality and Traffic datasets for longer horizons. These detailed observations confirm the effectiveness of the combined approach in improving forecasting accuracy, especially in longer prediction horizons. This suggests that the integration of these components can effectively tackle the inherent challenges in LTSF problems, making it a valuable strategy for enhancing time series forecasting models.

\begin{figure*}[t]
    \centering
    \includegraphics[width=\linewidth]{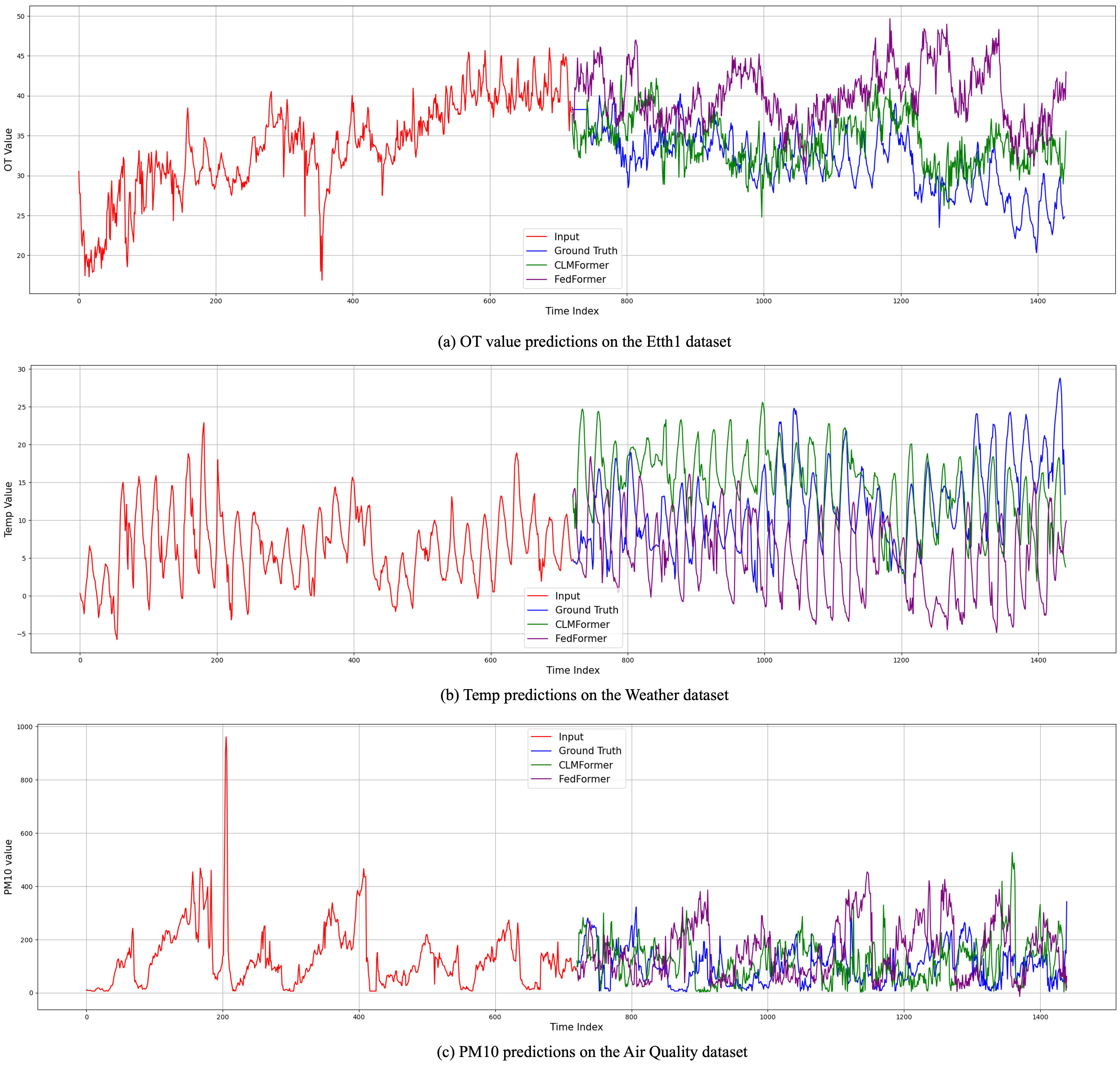}
    \caption{Samples of time series data from the ETTh1, Weather, and Air Quality datasets, along with predictions by FedFormer and our CLMFormer. We employ multivariate settings with prediction lengths of 720 and visualize the OT, temperature and PM10 value for the ETTh1, Weather and Air Quality datasets, respectively.}
    \label{fig:vis}
\end{figure*}

\mj{Table.\ref{ablation2} presents the findings from our second ablation study focused on evaluating the efficacy of different memory matrix initialization methods within our proposed model across three distinct datasets: Etth1, Etth2, and Weather. This study specifically compares the performance of our model using a seasonal memory matrix (denoted as "ours") against a baseline initialization approach employing an Identity matrix \cite{chen2020raio,memoryma2021}. The results delineated in the table underscore the superior performance of the Informer model augmented with our seasonal memory matrix across the majority of tasks and horizons, particularly highlighting its robustness in capturing long-term dependencies and complex temporal dynamics inherent in the datasets. For instance, in the Etth1 dataset, our approach consistently outperforms the Identity matrix initialization in both MSE and MAE metrics across all forecasting horizons, with significant improvements observed at longer horizons (336 and 720 predictions), which underscores the enhanced capacity of our model to retain and leverage long-term historical information for accurate future predictions. Moreover, the study also explores the efficacy of these memory matrix initialization methods within the FedFormer model, further substantiating the adaptability and effectiveness of our seasonal memory matrix in enhancing forecasting performance. Notably, the FedFormer model with our memory matrix initialization demonstrates remarkable improvements over the Identity matrix, especially in the Etth2 and Weather datasets, indicating the generalizability of our proposed method across different models and datasets. This ablation study clearly demonstrates that our seasonal memory matrix significantly outperforms the traditional Identity matrix in capturing long-term dependencies and intricate temporal patterns within time series data. By providing a more sophisticated mechanism for initializing the memory matrix, our approach enables the model to better understand and predict future values based on historical data, thereby setting a new benchmark for performance in time series forecasting tasks. This validation not only reinforces the value of our adaptations to the memory matrix but also highlights the potential for further innovations in enhancing the capabilities of time series forecasting models.}

\subsection{Case Study}

In this section, we embark on two case studies to conduct a qualitative evaluation of the predictive capabilities of our CLMFormer and FedFormer. Specifically, we present the input and ground truth instances with the predictions rendered by both FedFormer and CLMFormer over a forecast length of 720 intervals within the ETTh1, Weather and Air Quality datasets, as depicted in Figure.\ref{fig:vis}. Across both subplots, it is discernible that the predictions generated by our CLMFormer, represented in green, more closely adhere to the ground truth, delineated in blue, exhibiting similar trends in contrast to those of FedFormer. This convergence with the ground truth underscores the enhanced capability of CLMFormer to generate both extensive and high-fidelity predictions. Moreover, it becomes evident that the FedFormer’s predictions, illustrated in purple, tend to perpetuate the trends observed in the input data, adhering to a linear trajectory. In contrast, CLMFormer is adept at capturing non-linear relationships. We postulate that this is attributable to our proposed seasonal memory-driven forecasting modules, which empower the model to internalize seasonal patterns from the entire corpus of training data, thereby furnishing more accurate and elongated predictive insights.

%% file: table/tab2.tex
\begin{table}[ht]
\centering
\caption{Key parameters for backbones.}
\label{trans1}
\begin{tabular}{c|rrr} 
\toprule
Prediction Len. & \{24,48\} & \{168,336\} & \{720\}  \\ 
\hline\hline
$d_{model}$    & 1024    & 1024      & 1024    \\
$d_{ff}$         & 2048    & 2048      & 2048    \\
n\_heads        & 8       & 8         & 8       \\
Dropout         & 0.1     & 0.1       & 0.1     \\
Activation      & GELU    & GELU      & GELU    \\
Batch Size      & 32      & 8         & 4       \\
Enc.Layer no.   & 1       & 2         & 2       \\
Dec.Layer no.   & 1       & 1         & 1       \\
\bottomrule
\end{tabular}
\end{table}

\begin{table}[ht]
\centering
\caption{Key parameters for the seasonal memory unit.}
\label{rm1}
\begin{tabular}{c|ccc} 
\toprule
\multicolumn{1}{r|}{} & AutoFormer & Fedformer  & Informer  \\ 
\hline\hline
$d_{rm}$             & 1024     & 1024         & 1024       \\
Mem.slot.no.          & 1        & 1            & 1          \\
Mem.head.no.          & 2        & 2            & 4          \\
\bottomrule
\end{tabular}
\end{table}

%% file: table/tab3.tex
\begin{table*}[t]
\centering
\caption{The multivariate results comparison of our method with the SOTA Transformer-based LTSF models (three repetitions). The numbers in bold indicate the better-performing models within each Transformer-based LTSF model pair.}
\label{baselinetable}
\resizebox{\textwidth}{!}{\begin{tabular}{c|c|cc|cc|cc|cc|cc|cc} 
\hline
\multicolumn{2}{c|}{Methods}    & \multicolumn{2}{c|}{FedFormer+ours} & \multicolumn{2}{c|}{FedFormer}  & \multicolumn{2}{c|}{Informer+ours} & \multicolumn{2}{c|}{Informer}     & \multicolumn{2}{c|}{Transformer+ours} & \multicolumn{2}{c}{Transformer}    \\ 
\hline
\multicolumn{2}{c|}{Metric}    & MSE            & MAE                & MSE            & MAE            & MSE            & MAE               & MSE             & MAE             & MSE             & MAE                 & MSE             & MAE              \\ 
\hline
\multirow{5}{*}{Etth1}   & 24  & \textbf{0.304} & \textbf{0.370}     & 0.315          & 0.381          & \textbf{0.495} & \textbf{0.493}    & 0.550           & 0.541           & 0.474           & 0.493               & \textbf{0.456 } & \textbf{0.474 }  \\
                         & 48  & \textbf{0.335} & \textbf{0.381}     & 0.338          & 0.392          & \textbf{0.582} & \textbf{0.555}    & 0.644           & 0.606           & \textbf{0.585}  & \textbf{0.568}      & 0.608           & 0.584            \\
                         & 168 & \textbf{0.407} & \textbf{0.438}     & 0.420          & 0.443          & \textbf{0.981} & \textbf{0.771}    & 1.110           & 0.861           & \textbf{0.914}  & \textbf{0.750}      & 0.925           & 0.767            \\
                         & 336 & \textbf{0.451} & \textbf{0.454}     & 0.459          & 0.465          & \textbf{1.098} & \textbf{0.816}    & 1.237           & 0.893           & \textbf{1.001 } & \textbf{0.801 }     & 1.077           & 0.821            \\
                         & 720 & \textbf{0.467} & \textbf{0.489}     & 0.506          & 0.507          & \textbf{1.156} & \textbf{0.842}    & 1.475           & 0.993           & \textbf{1.164}  & \textbf{0.871}      & 1.182           & 0.872            \\ 
\hline
\multirow{5}{*}{Etth2}   & 24  & 0.224          & 0.316     & \textbf{0.220} & \textbf{0.312}          & 0.542          & 0.560             & \textbf{0.385 } & \textbf{0.467 } & 0.534           & 0.581               & \textbf{0.474 } & \textbf{0.522 }  \\
                         & 48  & \textbf{0.271} & \textbf{0.358}     & 0.284          & 0.47           & \textbf{1.190} & \textbf{0.873}    & 1.959           & 1.119           & \textbf{0.909}  & \textbf{0.761}      & 1.063           & 0.832            \\
                         & 168 & \textbf{0.405} & \textbf{0.410}     & 0.412          & 0.427          & \textbf{4.869} & \textbf{1.856}    & 5.957           & 2.036           & 4.951           & 1.807               & \textbf{4.067 } & \textbf{1.623 }  \\
                         & 336 & \textbf{0.481} & \textbf{0.469}     & 0.496          & 0.487          & 4.672          & 1.872             & \textbf{4.303 } & \textbf{1.780 } & 4.371           & 1.727               & \textbf{3.852 } & \textbf{1.613 }  \\
                         & 720 & \textbf{0.442} & \textbf{0.453}     & 0.463          & 0.474          & \textbf{3.358} & \textbf{1.569}    & 3.444           & 1.574           & 3.690           & 1.671               & \textbf{2.603 } & \textbf{1.369 }  \\ 
\hline
\multirow{5}{*}{Air}     & 24  & 0.590          & 0.411              & \textbf{0.588} & \textbf{0.412} & 0.542          & \textbf{0.374}    & \textbf{0.562 } & 0.388           & 0.560           & 0.383               & \textbf{0.552 } & \textbf{0.380 }  \\
                         & 48  & \textbf{0.694} & 0.459              & 0.686          & \textbf{0.457} & \textbf{0.650} & \textbf{0.435}    & 0.692           & 0.447           & \textbf{0.679}  & \textbf{0.440}      & 0.698           & 0.448            \\
                         & 168 & \textbf{0.805} & \textbf{0.504}     & 0.816          & 0.515          & \textbf{0.720} & \textbf{0.479}    & 0.883           & 0.518           & \textbf{0.780}  & 0.502               & 0.798           & \textbf{0.499}   \\
                         & 336 & \textbf{0.827} & \textbf{0.515}     & 0.834          & 0.518          & \textbf{0.773} & \textbf{0.505}    & 0.852           & 0.536           & \textbf{0.789}  & \textbf{0.501}      & 0.823           & 0.514            \\
                         & 720 & \textbf{0.885} & \textbf{0.526}     & 0.898          & 0.540          & \textbf{0.793} & \textbf{0.503}    & 0.861           & 0.534           & \textbf{0.840}  & \textbf{0.514}      & 0.859           & 0.524            \\ 
\hline
\multirow{5}{*}{Traffic} & 24  & 0.611          & 0.412              & \textbf{0.556} & \textbf{0.364} & \textbf{0.324} & \textbf{0.222}    & 0.381           & 0.264           & \textbf{0.337}  & \textbf{0.248}      & 0.411           & 0.282            \\
                         & 48  & 0.588          & 0.373              & \textbf{0.561} & \textbf{0.359} & \textbf{0.351} & \textbf{0.243}    & 0.440           & 0.300           & \textbf{0.364}  & \textbf{0.275}      & 0.486           & 0.336            \\
                         & 168 & \textbf{0.587} & \textbf{0.364}     & 0.612          & 0.381          & \textbf{0.405} & \textbf{0.300}    & 0.567           & 0.396           & \textbf{0.407}  & \textbf{0.300}      & 0.555           & 0.384            \\
                         & 336 & \textbf{0.584} & \textbf{0.350}     & 0.621          & 0.383          & \textbf{0.397} & \textbf{0.291}    & 0.572           & 0.399           & \textbf{0.413}  & \textbf{0.302}      & 0.570           & 0.400            \\
                         & 720 & \textbf{0.597} & \textbf{0.351}     & 0.626          & 0.382          & \textbf{0.468} & \textbf{0.330}    & 0.585           & 0.400           & \textbf{0.392}  & \textbf{0.293}      & 0.585           & 0.399            \\ 
\hline
\multirow{5}{*}{Weather} & 24  & \textbf{0.148} & \textbf{0.231}     & 0.162          & 0.247          & \textbf{0.253} & \textbf{0.338}    & 0.258           & 0.342           & \textbf{0.253}  & \textbf{0.340}      & 0.263           & 0.352            \\
                         & 48  & \textbf{0.189} & \textbf{0.274}     & 0.204          & 0.293          & \textbf{0.344} & \textbf{0.411}    & 0.354           & 0.418           & \textbf{0.338}  & \textbf{0.414}      & 0.348           & 0.419            \\
                         & 168 & \textbf{0.263} & \textbf{0.325}     & 0.284          & 0.347          & \textbf{0.507} & \textbf{0.520}    & 0.578           & 0.544           & \textbf{0.513}  & \textbf{0.512}      & 0.551           & 0.526            \\
                         & 336 & \textbf{0.305} & \textbf{0.356}     & 0.349          & 0.390          & \textbf{0.545} & \textbf{0.537}    & 0.591           & 0.553           & \textbf{0.512}  & \textbf{0.507}      & 0.584           & 0.547            \\
                         & 720 & \textbf{0.384} & \textbf{0.395}     & 0.413          & 0.428          & \textbf{0.593} & \textbf{0.567}    & 0.625           & 0.643           & \textbf{0.554}  & \textbf{0.530}      & 0.642           & 0.576 \\
                         \bottomrule
\end{tabular}}
\end{table*}

%% file: table/tab1.tex
\begin{table*}[t]
\centering
\caption{The univariate results comparison of our method with the Transformer-based LTSF models. The numbers in bold and red indicate the better-performing models within each Transformer-based LTSF model pair and compared to the InParformer. \mj{These improvements are statistically significant at $p = 0.05$.}}
\label{baselinetable-uni}
\resizebox{\textwidth}{!}{\begin{tabular}{c|c|cc|cc|cc|cc|cc} 
\toprule
\multicolumn{2}{c|}{Methods}              & \multicolumn{2}{c|}{FedFormer+ours}              & \multicolumn{2}{c|}{FedFormer}   & \multicolumn{2}{c|}{AutoFormer+ours}              & \multicolumn{2}{c|}{AutoFormer}  & \multicolumn{2}{c}{InParformer}  \\ 
\hline
\multicolumn{2}{c|}{Metric}               & MSE                             & MAE            & MSE             & MAE            & MSE             & MAE                             & MSE             & MAE            & MSE    & MAE                     \\ 
\hline
\multirow{4}{*}{\rotcell{Weather}}  & 96  & 0.0065                          & 0.062          & \textbf{0.0062} & 0.062          & 0.0111          & 0.083                           & \textbf{0.0110} & \textbf{0.081} & 0.0022 & 0.036                   \\
                                    & 192 & 0.0061                          & 0.062          & \textbf{0.0060} & 0.062          & \textbf{0.0074} & 0.067                           & 0.0075          & 0.067          & 0.0038 & 0.048                   \\
                                    & 336 & \textbf{0.0039}                 & \textbf{0.048} & 0.0041          & 0.050          & \textbf{0.0060} & \textbf{0.059}                  & 0.0063          & 0.062          & 0.0033 & 0.045                   \\
                                    & 720 & \textbf{0.0051}                 & \textbf{0.058} & 0.0055          & 0.059          & \textbf{0.0079} & \textbf{0.066}                  & 0.0085          & 0.070          & 0.0030 & 0.042                   \\ 
\hline
\multirow{4}{*}{\rotcell{ETTh2}}    & 96  & \textbf{0.126}                  & \textbf{0.270} & 0.128           & 0.271          & \textbf{0.150}  & \textbf{0.301}                  & 0.153           & 0.306          & 0.117  & 0.264                   \\
                                    & 193 & 0.188                           & \textbf{0.319} & \textbf{0.185}  & 0.330          & \textbf{0.200}  & \textbf{0.347}                  & 0.204           & 0.351          & 0.169  & 0.319                   \\
                                    & 336 & \textbf{0.221}                  & \textbf{0.364} & 0.231           & 0.378          & \textbf{0.238}  & \textbf{0.382}                  & 0.246           & 0.389          & 0.225  & 0.373                   \\
                                    & 720 & \textbf{0.267}                  & \textbf{0.406} & 0.278           & 0.420          & \textbf{0.252}  & \textbf{\textcolor{red}{0.398}} & 0.268           & 0.409          & 0.241  & 0.399                   \\ 
\hline
\multirow{4}{*}{\rotcell{Exchange}} & 96  & 0.155                           & \textbf{0.301} & \textbf{0.154}  & 0.304          & \textbf{0.239}  & 0.388                           & 0.241           & \textbf{0.387} & 0.105  & 0.247                   \\
                                    & 193 & \textbf{0.284}                  & 0.427          & 0.286           & \textbf{0.420} & \textbf{0.295}  & \textbf{0.366}                  & 0.300           & 0.369          & 0.207  & 0.360                   \\
                                    & 336 & \textbf{0.498}                  & \textbf{0.552} & 0.511           & 0.555          & \textbf{0.498}  & \textbf{0.517}                  & 0.509           & 0.524          & 0.400  & 0.498                   \\
                                    & 720 & \textbf{\textcolor{red}{1.126}} & \textbf{0.860} & 1.301           & 0.879          & \textbf{1.167}  & \textbf{\textcolor{red}{0.832}} & 1.260           & 0.867          & 1.172  & 0.836          \\
                                    \bottomrule
\end{tabular}}
\end{table*}

%% file: table/tab4.tex
\begin{table*}[t]
\caption{Ablation study for the decomposition of our approach, where p.t. is the progressive training scheme and mem refers to our memory approach (three repetitions).}
\label{ablation1}
\resizebox{\textwidth}{!}{
\begin{tabular}{lc|cccc|cccc|cccc}
\toprule
\multicolumn{2}{c|}{datasets}                                                                  & \multicolumn{4}{c|}{Eth1}                               & \multicolumn{4}{c|}{Air quality}                       & \multicolumn{4}{c}{Traffic}                              \\ \hline
\multicolumn{2}{c|}{prediction}                                                                         & 24             & 48             & 168            & 336            & 24             & 48             & 168            & 336            & 24             & 48             & 168            & 336            \\ \toprule
\multicolumn{1}{l|}{\multirow{2}{*}{Informer}}                                                    & MSE & 0.550          & 0.644          & 1.110          & 1.237          & 0.562          & 0.692          & 0.833          & 0.852          & 0.381          & 0.440          & 0.567          & 0.572          \\
\multicolumn{1}{l|}{}                                                                             & MAE & 0.541          & 0.606          & 0.861          & 0.893          & 0.388          & 0.447          & 0.518          & 0.536          & 0.264          & 0.300          & 0.396          & 0.399          \\ \hline
\multicolumn{1}{l|}{\multirow{2}{*}{\begin{tabular}[c]{@{}l@{}}Informer \\ + p.t.\end{tabular}}}  & MSE & 0.510          & 0.651          & 1.021          & 1.136          & 0.583          & 0.690          & 0.813          & 0.848          & 0.377          & 0.431          & 0.564          & 0.571          \\
\multicolumn{1}{l|}{}                                                                             & MAE & 0.514          & 0.608          & 0.820          & 0.820          & 0.387          & 0.447          & 0.524          & 0.535          & 0.262          & 0.290          & 0.389          & 0.398          \\ \hline
\multicolumn{1}{l|}{\multirow{2}{*}{\begin{tabular}[c]{@{}l@{}}Informer \\ + mem\end{tabular}}}   & MSE & 0.510          & 0.588          & 1.009          & 1.161          & 0.569          & 0.680          & 0.769          & 0.794          & 0.331          & 0.352          & 0.411          & 0.417          \\
\multicolumn{1}{l|}{}                                                                             & MAE & 0.506          & 0.556          & 0.810          & 0.827          & 0.386          & 0.438          & 0.496          & 0.516          & 0.228          & 0.249          & 0.302          & 0.306          \\ \hline
\multicolumn{1}{l|}{\multirow{2}{*}{\begin{tabular}[c]{@{}l@{}}Informer \\ + ours\end{tabular}}}  & MSE & \textbf{0.495} & \textbf{0.582} & \textbf{0.981} & \textbf{1.098} & \textbf{0.542} & \textbf{0.650} & \textbf{0.720} & \textbf{0.773} & \textbf{0.324} & \textbf{0.351} & \textbf{0.405} & \textbf{0.397} \\
\multicolumn{1}{l|}{}                                                                             & MAE & \textbf{0.493} & \textbf{0.555} & \textbf{0.771} & \textbf{0.816} & \textbf{0.374} & \textbf{0.435} & \textbf{0.479} & \textbf{0.505} & \textbf{0.222} & \textbf{0.243} & \textbf{0.300} & \textbf{0.291} \\ \toprule
\multicolumn{1}{l|}{\multirow{2}{*}{FedFormer}}                                                   & MSE & 0.315          & 0.338          & 0.420          & 0.459          & \textbf{0.588} & \textbf{0.686} & 0.816          & 0.834          & \textbf{0.556} & \textbf{0.561} & 0.612          & 0.621          \\
\multicolumn{1}{l|}{}                                                                             & MAE & 0.381          & 0.392          & 0.443          & 0.465          & 0.412          & \textbf{0.457} & 0.515          & 0.518          & \textbf{0.364} & \textbf{0.359} & 0.381          & 0.383          \\ \hline
\multicolumn{1}{l|}{\multirow{2}{*}{\begin{tabular}[c]{@{}l@{}}FedFormer \\ + p.t.\end{tabular}}} & MSE & 0.314          & 0.336          & 0.415          & 0.454          & 0.593          & 0.694          & 0.818          & 0.833          & 0.608          & 0.572          & 0.603          & 0.601          \\
\multicolumn{1}{l|}{}                                                                             & MAE & 0.377          & 0.383          & 0.440          & 0.459          & 0.412          & 0.459          & 0.513          & 0.518          & 0.392          & 0.366          & 0.378          & 0.374          \\ \hline
\multicolumn{1}{l|}{\multirow{2}{*}{\begin{tabular}[c]{@{}l@{}}FedFormer \\ + mem\end{tabular}}}  & MSE & 0.308          & 0.336          & 0.411          & \textbf{0.450} & 0.590          & 0.689          & 0.812          & 0.830          & 0.575          & 0.567          & 0.601          & 0.590          \\
\multicolumn{1}{l|}{}                                                                             & MAE & 0.372          & 0.385          & 0.446          & 0.461          & 0.414          & \textbf{0.457} & 0.509          & \textbf{0.515} & 0.374          & 0.362          & 0.368          & 0.359          \\ \hline
\multicolumn{1}{l|}{\multirow{2}{*}{\begin{tabular}[c]{@{}l@{}}FedFormer \\ + ours\end{tabular}}} & MSE & \textbf{0.304} & \textbf{0.335} & \textbf{0.407} & 0.451          & 0.590          & 0.694          & \textbf{0.805} & \textbf{0.827} & 0.611          & 0.588          & \textbf{0.587} & \textbf{0.584} \\
\multicolumn{1}{l|}{}                                                                             & MAE & \textbf{0.370} & \textbf{0.381} & \textbf{0.438} & \textbf{0.454} & \textbf{0.411} & 0.459          & \textbf{0.504} & \textbf{0.515} & 0.412          & 0.373          & \textbf{0.364} & \textbf{0.350} \\ \toprule
\end{tabular}}
\end{table*}

%% file: table/tab5.tex
\begin{table*}[t]
\centering
\caption{Ablation study of different memory matrix initialization methods (three repetitions).}
\label{ablation2} 
\resizebox{\textwidth}{!}{
\begin{tabular}{lc|cccc|cccc|cccc}
\toprule
\multicolumn{2}{c|}{datasets}                                                                                         & \multicolumn{4}{c|}{Etth1}                                        & \multicolumn{4}{c|}{Etth2}                                        & \multicolumn{4}{c}{Weather}                                       \\ \hline
\multicolumn{2}{c|}{prediction}                                                                                       & 24             & 48             & 336            & 720            & 24             & 48             & 168            & 336            & 24             & 48             & 168            & 336            \\ \toprule
\multicolumn{1}{l|}{\multirow{2}{*}{\begin{tabular}[c]{@{}l@{}}Informer+mem\\ (ours)\end{tabular}}}             & MSE & \textbf{0.510} & 0.588          & \textbf{1.009} & \textbf{1.161} & \textbf{0.553} & \textbf{1.204} & \textbf{5.069} & \textbf{4.883} & \textbf{0.243} & \textbf{0.350} & \textbf{0.519} & \textbf{0.556} \\
\multicolumn{1}{l|}{}                                                                                           & MAE & \textbf{0.506} & 0.556          & \textbf{0.810} & \textbf{0.843} & \textbf{0.578} & \textbf{0.879} & \textbf{2.043} & \textbf{1.869} & \textbf{0.328} & \textbf{0.411} & \textbf{0.525} & \textbf{0.544} \\ \hline
\multicolumn{1}{l|}{\multirow{2}{*}{\begin{tabular}[c]{@{}l@{}}Informer+mem\\ (Identity matrix)\end{tabular}}}  & MSE & 0.537          & \textbf{0.571} & 1.168          & 1.224          & 0.592          & 1.242          & 6.454          & 5.034          & 0.254          & 0.353          & 0.521          & 0.567          \\
\multicolumn{1}{l|}{}                                                                                           & MAE & 0.528          & \textbf{0.546} & 0.848          & 0.894          & 0.602          & 0.888          & 2.251          & 1.943          & 0.335          & 0.412          & 0.530          & 0.549          \\ \toprule
\multicolumn{1}{l|}{\multirow{2}{*}{\begin{tabular}[c]{@{}l@{}}FedFormer+mem\\ (ours)\end{tabular}}}            & MSE & \textbf{0.308} & \textbf{0.336} & \textbf{0.454} & \textbf{0.476} & 0.222          & \textbf{0.280} & \textbf{0.409} & \textbf{0.488} & \textbf{0.153} & \textbf{0.194} & \textbf{0.268} & \textbf{0.311} \\
\multicolumn{1}{l|}{}                                                                                           & MAE & \textbf{0.377} & \textbf{0.383} & \textbf{0.459} & \textbf{0.463} & \textbf{0.314} & \textbf{0.381} & \textbf{0.416} & \textbf{0.471} & \textbf{0.239} & \textbf{0.280} & \textbf{0.332} & \textbf{0.365} \\ \hline
\multicolumn{1}{l|}{\multirow{2}{*}{\begin{tabular}[c]{@{}l@{}}FedFormer+mem\\ (Identity matrix)\end{tabular}}} & MSE & 0.310          & 0.352          & 0.467          & 0.488          & \textbf{0.218} & 0.283          & 0.417          & 0.503          & 0.167          & 0.211          & 0.276          & 0.325          \\
\multicolumn{1}{l|}{}                                                                                           & MAE & 0.383          & 0.394          & 0.471          & 0.480          & 0.318          & 0.390          & 0.422          & 0.484          & 0.246          & 0.292          & 0.357          & 0.381          \\ \toprule
\end{tabular}}
\end{table*}